\documentclass[acmsmall,screen]{acmart}
\renewcommand\footnotetextcopyrightpermission[1]{} % removes footnote with conference information in first column
\AtBeginDocument{%
  \providecommand\BibTeX{{%
    \normalfont B\kern-0.5em{\scshape i\kern-0.25em b}\kern-0.8em\TeX}}}

\setcopyright{acmcopyright}
\copyrightyear{2018}
\acmYear{2018}
\acmDOI{10.1145/1122445.1122456}

\acmJournal{JACM}
\acmVolume{37}
\acmNumber{4}
\acmArticle{111}
\acmMonth{8}

\usepackage{indentfirst}
\usepackage{color}
\usepackage{url}

\begin{document}

\title{Aesthetic Language Guidance Generation of Images Using Attribute Comparison}

\author{Xin Jin}
\email{jinxinbesti@foxmail.com}
\affiliation{%
	\institution{Beijing Electronic Science and Technology Institute}
	\city{Beijing}
	\state{Beijing}
	\country{China}
}

\author{Qiang Deng}
\email{1352110584@qq.com}
\affiliation{%
	\institution{Beijing Electronic Science and Technology Institute}
	\city{Beijing}
	\state{Beijing}
	\country{China}
}

\author{Jianwen Lv}
\email{513415184@qq.com}
\affiliation{%
	\institution{Beijing Electronic Science and Technology Institute}
	\city{Beijing}
	\state{Beijing}
	\country{China}
}

\author{Heng Huang}
\email{1264123696@qq.com}
\affiliation{%
	\institution{Beijing Electronic Science and Technology Institute}
	\city{Beijing}
	\state{Beijing}
	\country{China}
}

\author{Hao Lou}
\email{452392771@qq.com}
\affiliation{%
	\institution{Beijing Electronic Science and Technology Institute}
	\city{Beijing}
	\state{Beijing}
	\country{China}
}

\author{Chaoen Xiao$^{*}$}
\email{xcecd@qq.com}
\thanks{*Corresponding authors}
\affiliation{%
	\institution{Beijing Electronic Science and Technology Institute}
	\city{Beijing}
	\state{Beijing}
	\country{China}
}


%%\begin{figure}[ht]
%%	\centering
%%	\includegraphics[width=\linewidth]{fig1.pdf}
%%	\caption{The example of three scores of the general aesthetic attributes}
%%\end{figure}

\begin{abstract}
With the vigorous development of mobile photography technology, major mobile phone manufacturers are scrambling to improve the shooting ability of equipments and the photo beautification algorithm of software. However, the improvement of intelligent equipments and algorithms cannot replace human subjective photography technology. In this paper, we propose the aesthetic language guidance of image (ALG). We divide ALG into ALG-T and ALG-I according to whether the guiding rules are based on photography templates or guidance images. Whether it is ALG-T or ALG-I, we guide photography from three attributes of color, lighting and composition of the images. The differences of the three attributes between the input images and the photography templates or the guidance images are described in natural language, which is aesthetic natural language guidance (ALG). Also, because of the differences in lighting and composition between landscape images and portrait images, we divide the input images into landscape images and portrait images. Both ALG-T and ALG-I conduct aesthetic language guidance respectively for the two types of input images (landscape images and portrait images).
\end{abstract}




\maketitle

\section{Introduction}
With the popularity and updating of the digital equipments, more and more people are focusing on image aesthetics. Especially nowadays the popularity of smart products has made people have higher and higher requirements for photographic quality. Nowadays, many smart product manufacturers are accelerating the hardware research and development of shooting equipments, and there are many beautifying softwares that are constantly improving the performance of beautifying algorithms. But they are more of beautification of the pictures taken, without any photography guidance to the photographer. 

We propose aesthetic language guidance of image (ALG) that can guide photographers when taking pictures, and we also provide two different approaches. If the guidance rules are based on the photography templates, it is ALG-T; if the guidance rules are based on the guidance images, it is ALG-I. Whether it is a photography template or a guidance image, aesthetic language of the image is guided from the three attributes of color, composition, and lighting. Because landscape images and portrait images are different in calculating composition and lighting attributes, these two attributes need to be calculated respectively for these two types of images, while the method for calculating the color attribute is the same for both types of images, whether in ALG-T or ALG-I.

\begin{figure}[htbp]
	\centering
	\includegraphics[height=3cm]{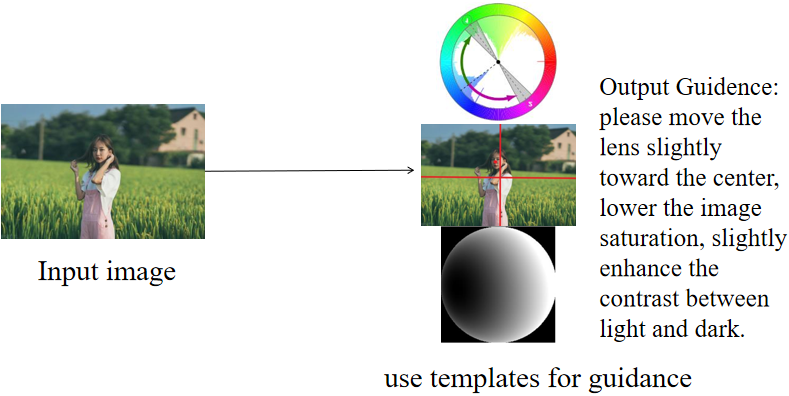}
	\caption{aesthetic language guidance based on photography templates (ALG-T)}
	\label{model}
\end{figure}

Figure \ref{model} is an example of using the ALG-T method to guide a portrait. As shown in Figure \ref{model}, the method of using templates for guidance is to calculate the preset image feature values, comparing it with the different values of the templates, and output it in natural language.
\begin{figure}[htbp]
	\centering
	\includegraphics[height=2.5cm]{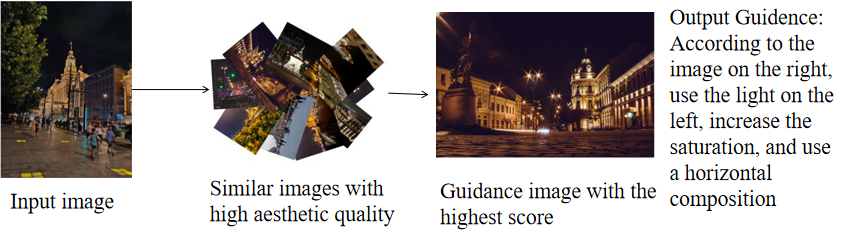}
	\caption{aesthetic language guidance based on guidance images (ALG-I)}
	\label{landscape}
\end{figure}

Figure \ref{landscape} is an example of using the ALG-I method to guide a landscape image. As shown in \ref{landscape}, our aesthetic guidance idea is to use a search model to search for any image that is similar to the input image and has a high aesthetic score, and pick the image with the highest aesthetic score as the guidance image. Then, depending on whether the input image is a landscape image or a portrait image, three aesthetic attributesare calculated. Finally, compare the differences of the three attributes between the input image and the guidance image, formulate appropriate rules, and describe them in natural language.

In conclusion, Main contributions of this paper include:  

1.To the best of our knowledge, this is the first work that proposes Aesthetic Language Guidance (ALG) for photography.

2.We propose ALG via photography templates (ALG-T) for landscape and portrait photography.

3.We propose ALG based on guidance images (ALG-I) for landscape and portrait photography. 

%%\begin{figure}[ht]
%%	\centering
%%	\includegraphics[width=\linewidth]{fig1.pdf}
%%	\caption{The example of three scores of the general aesthetic attributes}
%%\end{figure}
\section{Related work}
Image aesthetic assessment is chiefly based on the aesthetic perception of image characteristics and the law of photography. Recent studies \cite{lu2014rapid}, \cite{lu2015deep}, \cite{kong2016photo}, \cite{mai2016composition} have revealed that using data-driven methods is more efficient because the quantity of training data available has risen from hundreds of images to millions of images. Researchers first explored the global characteristics. Data et al.\cite{datta2006studying} and Ke et al.\cite{ke2006design} is one of the first people to transform the aesthetic perception of images into binary classification. Data et al.\cite{datta2006studying} combined the low-level features and high-level features commonly in image retrieval and trained support vector machine classifier for binary classification of image aesthetic quality.

A simple method of using depth learning method is to use the general depth characteristics obtained from other assignments, and train a new classifier on aesthetic classification tasks. Kong et al.\cite{kong2016photo} raised to assist learning aesthetic features by pairwise sorting of image pairs, image attributes and content information. Specifically, a connected structure with image pair as input is adopted, in which the two basic networks of the connected structure are configured with alexnet (excluding the fc8 of 1000 class classification layer from alexnet). Recently, Lu et al.\cite{lu2020deep} have found that using deep convolutional networks in the field of image retrieval can better construct the similarity (visually or semantically) between pairs of images. As we all  know, similarity construction of images is the premise of ratio. Jun-Tae Lee et al.\cite{lee2019image}  through two steps, constructing a ratio matrix of multiple sets of pictures and then using some selected images as the reference images, to predict aesthetic scores. Kao et al.\cite{kao2016visual} proposed multi-task learning of image aesthetics. Yan et al.\cite{yan2015change}, \cite{yan2013learning} aimed to explain the content removed and transformed by clipping itself, and try to contain the impact of the beginning composition of the initial image on the ending composition of the clipped image.

\section{ALG Based on Photography Templates (ALG-T)}
\subsection{Aesthetic Color Guidelines}
\subsubsection{Color Palettes of Images Based on Color Harmony}
\ 
\newline
\indent Most human perceptions of images are from color. Although our perceptions of color depend on environments and cultures, we are influenced by color harmony instead of mechanically evaluating images. The degree of color harmony in images not only depends on composition of various colors and colors’ proportion, but also depends on the relative positions of different colors in color space. Nowadays, there is no equation to define a harmonic set. However, there is a common sense in defining the harmonic set: the degree of color harmony is calculated by various colors’ format and relation in color space.
\begin{figure}
	\centering
	\includegraphics[height=3cm]{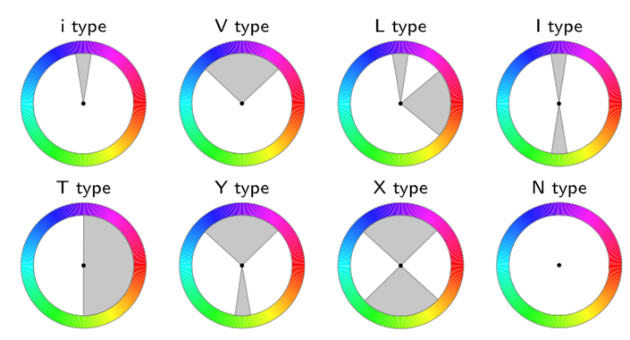}
	\caption{color harmony palette in the color rings}
	\label{palettes}
\end{figure}
Itten \cite{1961The}proposed a series of new color rings in Figure \ref{palettes}. If the color in the image is in the shadow area, we believe the image color is harmonic. Color harmony palette can be rotated at any angle. Itten described the degree of color harmony. Itten’s color harmony theory is based on the relative position of colors in the color ring. For example, matching three basic colors, ultramarine blue, deep red and yellow, can generate a 12-color color ring. If two colors are symmetric in color space, that can be called color harmony. He also realized that three-color harmony can generate any triangle; four-color harmony can generate any rectangle; five-color harmony can generate any pentagon. Based on Itten’s schemes, by combining several types of color tones and its distributions, Matsuda invented an 80-color scheme. Those schemes were used to both evaluate the degree of color harmony and design colors. Our color harmony method is also based on those schemes.

Matsuda proposed color harmony that was well accepted by the field of color of application. Figure \ref{palettes} shows that the color tone space, in HSV (Hue, Saturation, Value), defines eight harmonic types. Every color palette can be space distribution of color harmony palettes. If the color tone extracting from the image is in the shadow area of one palette, then, we will harmonize the image based on this palette. The geometric distributions of those color space are called eight color palettes. However, those color palettes only defined radial relationships of color ring instead of the given and continuous color space. In the other word, all color palettes can rotate at any angle; The palette regulates the relatively geometric distributions between color space. Color harmony palettes may consist of tones with the same color (type i, V and T), complementary colors (type i, Y, X) , or more complicated combinations. 

For any input image, we can fit color harmony palette Tm and HSV color tone histogram extracted from the input image. By the distance between the histogram and eight palettes, we can make sure the most suitable color palette for the input image. The palette Tm and the relative direction $\alpha$ f it can define a color harmony scheme. We use $(m,\alpha)$ to represent the scheme. For the given scheme $F(X,(m,\alpha))$, we can define a function $F(X,(m,\alpha))$, which measures the degree of harmony between image x and color harmony palette $(m,\alpha)$, like equation:
\begin{align} \label{equ_1}
	F\left( {X,\left( {m,\alpha } \right)} \right) = \sum\limits_{p \in X} {\left\| {H\left( p \right) - {E_{{T_m}\left( \alpha  \right)}}\left( p \right)} \right\| \cdot S\left( p \right)} 
\end{align}
In equation \ref{equ_1}, H is hue channel, and S is saturation channel; the distance of color tone represents the arc distance in the color ring. If HSV color tone histograms are in the shadow area of Tm, the distance of the input image and the corresponding palette is zero.
\begin{align} \label{equ_2}
	M\left( {X,{T_m}} \right) = \left( {m,{\alpha _0}} \right),{\alpha _0} = \mathop {\arg \min }\limits_\alpha  F\left( {X,\left( {m,\alpha } \right)} \right) 
\end{align}
For the given image X and palette Tm, the best color harmony scheme for X is the angle $\alpha\in[0,2\pi]$ meeting the minimum of equation \ref{equ_1}, shown as equation \ref{equ_2}.
\begin{align} \label{equ_3}
	B\left( X \right) = \left( {{m_0},{\alpha _0}} \right),{m_0} = \mathop {\arg \min }\limits_m F\left( {X,M\left( {X,{T_m}} \right)} \right) 
\end{align}
For the given image X, the best color harmony scheme is the palette meeting the minimum of equation \ref{equ_2} in all possible palettes, shown as equation \ref{equ_3}.

\subsubsection{Aesthetic Color Guidelines Based on Color Palettes}
\ 
\newline
\indent In the 3.1 section, eight color palettes were shown, as well as a color harmonizing algorithm that relies on color harmony. However, this naive definition does not take into account the spatial coherence between image pixels, which can lead to regions of discord due to "segmenting" adjacent regions of the image. 

To handle this situation, we use a segmentation method similar to Boykov, using an image cutting optimization technique. Specify the compressed fan for each pixel, and to ensure that the colors of the relatively continuous areas of the image are not compressed discontinuously, then we can first classify each pixel of the image. When there are two fans, the pixel's label is 0 or 1, indicating which area should be divided into.

With this technique of fine-tuning colors, we use it as the color guidance in the aesthetic guidance of images. The input image can be processed using the methods mentioned before to improve the overall color of the input image.

After we have the original image and its processed image, we can compare the richness of the colors. The standard we use is the color richness index defined by Hasler and Süsstrunk \cite{2003Measuring}, which is divided into 7 levels:

1. Not colorful

2. Slightly colorful

3. Moderately colorful

4. Averagely colorful

5. Very (Quite colorful)

6. Highly colorful

7. Extremely colorful
\begin{figure}
	\centering
	\includegraphics[height=3cm]{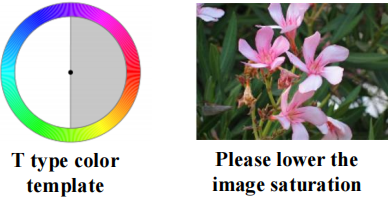}
	\caption{Aesthetic color guidance result}
	\label{color}
\end{figure}

Then by comparing the color richness of the two images, we can use language to guide whether our image should be more vivid or frosty, and a example is shown in Figure \ref{color}. 

\subsection{Aesthetic Lighting Guidelines for Landscape Images}
\begin{figure}[htbp]
	\centering
	\includegraphics[height=3cm]{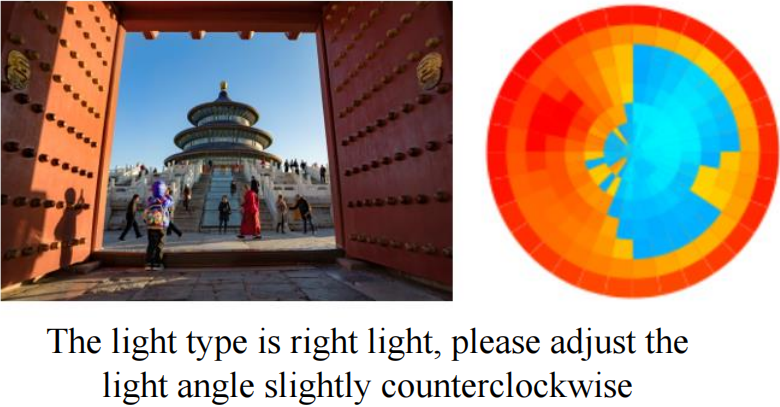}
	\caption{Lighting Guidance for Landscape Image}
	\label{lightingforl}
\end{figure}
This paper uses a deep learning-based technique that can estimate the lighting distribution of outdoor images\cite{2001Earthshine}\cite{L2013Adding}\cite{2004High}. Through a specially designed CNN model and outdoor panorama dataset. In these panoramas, an outdoor lighting model is applied to the sky to obtain a set of parameters (including the location of sunlight exposure, weather conditions such as clouds and haze, and camera parameters), and then a limited field of view is extracted from the panorama images, and train a CNN using the light parameters from these input images. Given a test image, the network can be used to infer lighting parameters, which in turn can be used to reconstruct outdoor light distribution maps. Lighting plays a pivotal role in photography and plays a vital role in determining the appearance of objects. Reverse analysis of light sources through images plays an important guiding role in adjusting the position of light sources in photography.
\begin{table}
	\begin{center}
		\caption{Sky Lighting Model Parameters}
		\label{structures}
		\begin{tabular}{lll}
			\hline\noalign{\smallskip}
			Layer & Stride & Resolution\\
			\noalign{\smallskip}
			\hline
			\noalign{\smallskip}
			Input & {} & 320*240\\
			Conv7-64 & {2} & 160*120\\
			Conv5-128 & {2} & 80*60\\
			Conv3-256 & {2} & 40*30\\
			Conv3-256 & {1} & 40*30\\
			Conv3-256 & {2} & 20*15\\
			Conv3-256 & {1} & 20*15\\
			Conv3-256 & {2} & 10*8\\
			& {FC-2048} & \\
			FC-160 LogSoftMax & {} & FC-5 Linear\\
			Output: sun position Distribution s & {} & Output: sky and camera parameters q\\
			\hline
		\end{tabular}
	\end{center}
\end{table}

Using the sky illumination model\cite{2012Recognizing}\cite{2012An}, the illumination intensity distribution of the image in the horizontal direction can be calculated. On this basis, this paper performs discretization processing and visual display, so that the horizontal distribution of light can be displayed in the form of a ring, as shown in Figure \ref{lightingforl}. The CNN structure used in this paper is shown in Table \ref{structures}: after 7 convolutional layers, a fully-connected layer is finally used to connect two heads of the model: one for regressing the sun position, and the other for regressing the sky and camera parameters.

The illumination circle obtained in Figure \ref{lightingforl} is divided into eight equal parts, corresponding to eight illumination directions, namely front light, back light, left light, right light, left front light, right front light, left back light, right back light. The center of each range of radians is used as its illumination direction. But a new problem will arise at this time. The illumination angle of the maximum light intensity of the entire image does not necessarily appear in the above eight directions. So by calculating the differences between the angle with the largest light intensity and the angle corresponding to the previously determined light type, the result is converted into natural language, which is the method to guide our landscape lighting. The result is shown in Figure \ref{lightingforl}.

\subsection{Aesthetic Lighting Guidelines for Portrait Images}
\begin{figure}
	\centering
	\includegraphics[height=3cm]{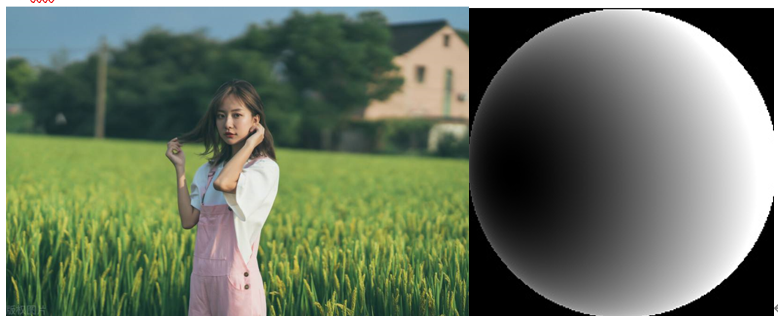}
	\caption{Spherical harmonics generated by portrait and face illumination}
	\label{nine}
\end{figure}
Traditional methods of portrait image illumination reconstruction based on physical illumination models need to solve rendering problems and estimate face geometry, reflectivity and illumination. However, the inaccurate estimation of face components can lead to strong artifacts in heavy lighting, resulting in suboptimal results. In this work, we apply a physically-based portrait relighting method to generate a large-scale, high-quality portrait lighting dataset (DPR). This dataset is then used to train a deep convolutional neural network (CNN) to generate a reshot portrait image. The model used in this paper is the structure of an hourglass network \cite{2016Stacked}. It has encoder and decoder sections. Four skip connections are used to connect features of different scales in the encoder part to features of corresponding scales in the decoder part. Using the above calculated portrait, the nine-dimensional spherical harmonic function can be used to characterize the illumination of the face. The original image and the calculated spherical harmonic function are shown in Figure \ref{nine}.

For the spherical harmonic function of face illumination generated in Figure \ref{nine}, in order to simplify the calculation, it can be projected in two dimensions, and the plane coordinate of the strongest illumination value is taken as its illumination center. And define the corresponding coordinate values of three light centers here: 

Rembrandt light: [82,172]

butterfly light: [63,127]

lower light: [191,127]

For any portrait image, light spherical harmonic function can be used to calculate the plane distance between its illumination center and the above three points. And Which coordinate point is the closest, then we think that the light type of the portrait image belongs to the type corresponding to this coordinate point. Rembrandt-style lighting technology can divide the light and shadow of the subject's face into two parts with obvious contrast, and make the two parts of the face light look different. If you use undifferentiated overlay lighting, the lighting details of the subject's face will be too convergent. So in most cases, Rembrandt lighting can be used to represent the lighting conditions of most parts of the face.
\begin{figure}
	\centering
	\includegraphics[height=5cm]{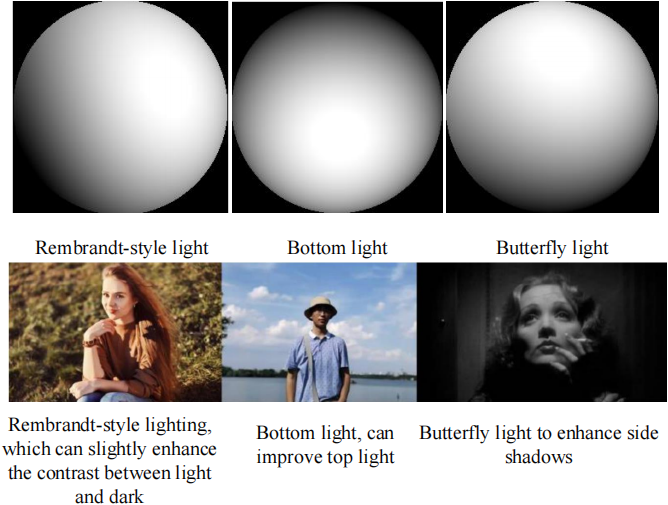}
	\caption{Portrait lighting guidance}
	\label{portraitlighting}
\end{figure}

Using spherical harmonics\cite{2019EfficientNet}\cite{2016Stacked}, we can calculate the degree of similarity between face lighting and Rembrandt lighting, butterfly lighting, and bottom lighting. If it is higher than 0.9, we consider it to be close. Otherwise, the lighting characteristics of such lighting should be strengthened. The three portrait lighting features are preset in advance, and the results are shown in Figure \ref{portraitlighting}.

\subsection{Aesthetic Composition Guidelines for Landscape Images}
\indent The semantic line labeling algorithm \cite{2016Richer}\cite{2016Holistically}\cite{1962Method}\cite{2020BSP}used in this paper is based on the deep Huffman transform semantic line detection. In order to distinguish the classification of the composition, an excellent model is needed to detect the semantic line of the image. The meaningful line structure in the image can divide the image to a certain extent, which is the semantic line.
\begin{figure}
	\centering
	\includegraphics[height=2cm]{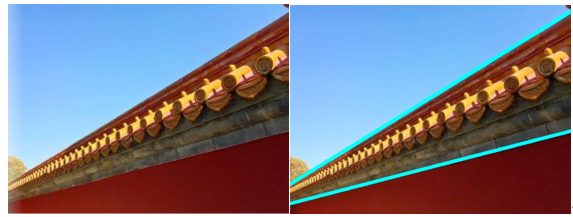}
	\caption{The calculation result of semantic line display}
	\label{semantic}
\end{figure}
On this basis, the standard input size of the image of the model used in this paper is unified to 640*426, so that it can calculate any image and describe it with the original image. Figure \ref{semantic} is an example of semantic line computation. 

Converting the calculated semantic line to the parameter type, the four coordinate points can be calculated as: ([0,363], [639,15]), ([0,402], [639,256]). 

Each of the two coordinate points corresponds to a semantic line, and the value of each coordinate corresponds to a pixel point of 640*426 in the image. The coordinate system is that the upper left corner of the image is the origin, the downward direction is the positive direction of the y-axis, and the right direction is the positive direction of the x-axis.
\begin{figure}
	\centering
	\includegraphics[height=3cm]{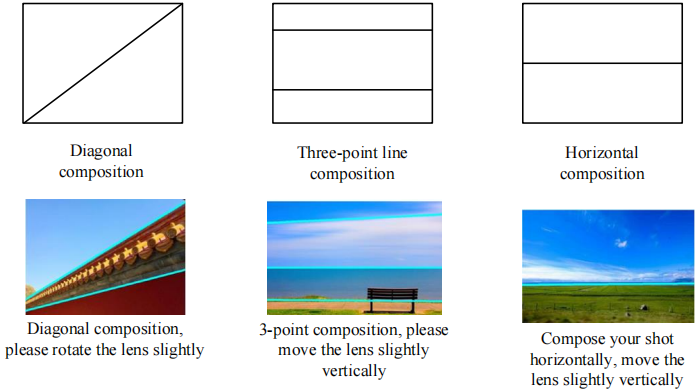}
	\caption{Landscape composition guidance}
	\label{lancomposition}
\end{figure}
Next we define several composition templates and set the parameters of the shooting guide. They are diagonal composition, thirds composition, and horizontal composition. By defining the positional relationship and angle between the semantic lines, the types of composition can be divided and guided to adjust the angle and parameters. The result is similar to Figure \ref{lancomposition}.

\subsection{Aesthetic Composition Guidelines for Portrait Images}
\indent In order to accurately identify humans from images and calibrate their positions, this paper uses a new object detection model named EfficientDet, which can always be achieved under the constraints of small-scale datasets and high real-time computing power requirements with better efficiency than existing models. EfficientNet \cite{2019EfficientNet} is adopted as the backbone network, BiFPN is adopted as the feature network, and a shared class/box prediction network is adopted.
\begin{figure}
	\centering
	\includegraphics[height=3cm]{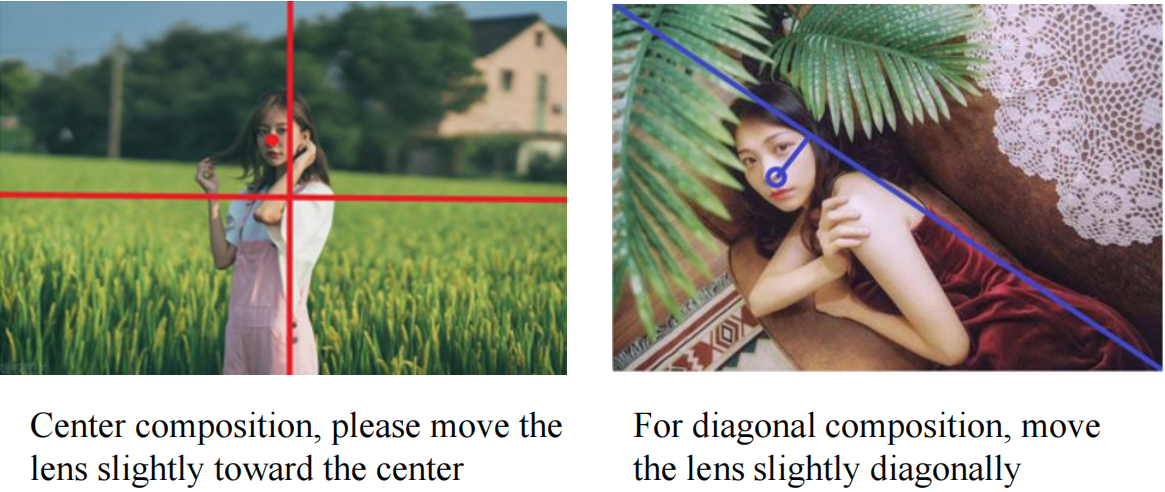}
	\caption{Portrait composition guidance}
	\label{porcomposition}
\end{figure}

By comparing the distance between the extracted facial feature points and the feature points of different compositions, as a basis for judging the type of composition, the result is shown in Figure \ref{porcomposition}.

\section{ALG Based on Guidance Images (ALG-I)}
\subsection{Image Search System}
OpenAI introduced a neural network called CLIP which efficiently learns visual concepts from natural language supervision. CLIP stands for Contrastive Language–Image Pre-training. For our use-case, we need to understand that CLIP is trained with a large set of images with their corresponding captions. So it learned to predict which image caption (text) matches closely with which image when a similarity metric (e.g.: Cosine) is applied on encodings of both image and text. So after training, we can give a random image and find cosine similarity of that image in the vector space with two vectors of phrases “Is this photo of a dog?”, “Is this photo of a cat?” and see which one has the highest similarity to find the class of the image. So in a way, it has zero-shot classification capabilities like GPT-2 and GPT-3. It is precise because of the powerful generalization ability of the clip model that we can easily classify images. In other words, image features and text vectors describing images can be converted to each other. Using this feature, we can calculate the feature value of each photo in the collected image database in advance, which can realize fast searching. For any image, we can find a batch of similar images through a search module. Through some image quality evaluation algorithms, we can score the searched images, and the image with the highest value can be used as our guidance image.
\begin{figure}
	\centering
	\includegraphics[height=3cm]{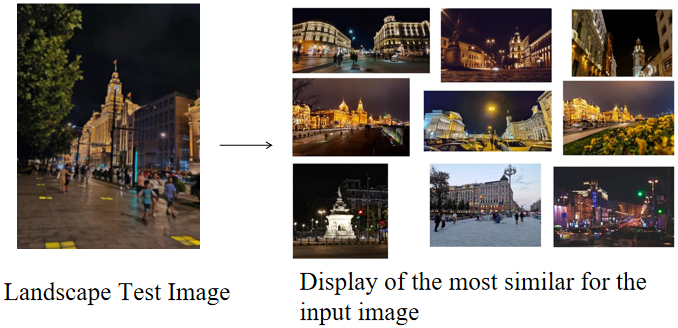}
	\caption{Image Search System for The Test Landscape Image}
	\label{input}
\end{figure}

The image on the left in Figure \ref{input} is the input image that needs to be searched for similarity. The results are displayed as the top nine most similar images, as shown in the image on the right in Figure \ref{input}.

\subsection{Aesthetic Landscape Images Guidance By ALG-I}
\begin{figure}[h!]
	\centering
	\includegraphics[height=3cm]{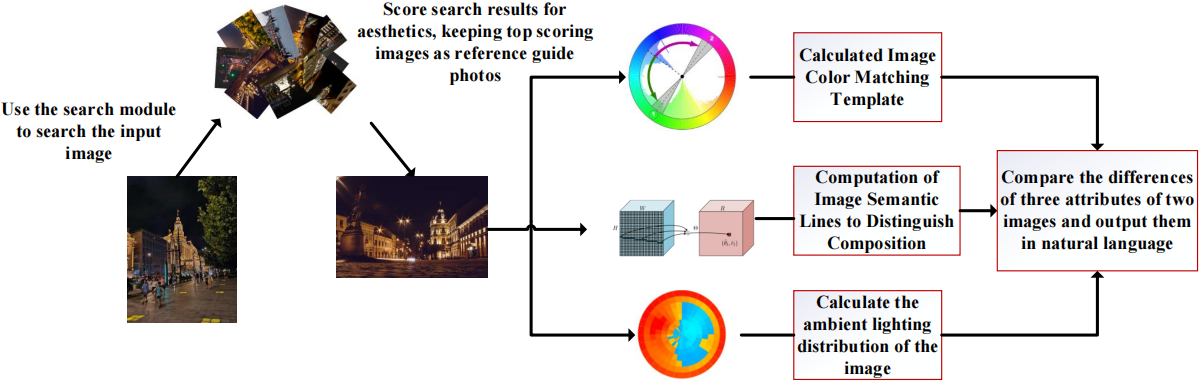}
	\caption{Landscape Aesthetic Image Guidance process}
	\label{landALGI}
\end{figure}

The basic idea of guidances for aesthetic landscape images is that for any landscape image, using our image search system in the 4.1, we can search for images similar to the provided landscape image. They are then scored using a specific aesthetic scoring engine, with the highest scoring landscape image selected as the guidance image. Now we have a image that needs guidance, and a guidance image, by comparing the differences in the three attributes of lighting, composition, and color between the two images, and compiling some rules to describe the differences in the attributes between the two images. The overall process is shown in Figure \ref{landALGI}.
\begin{figure}
	\centering
	\includegraphics[height=3cm]{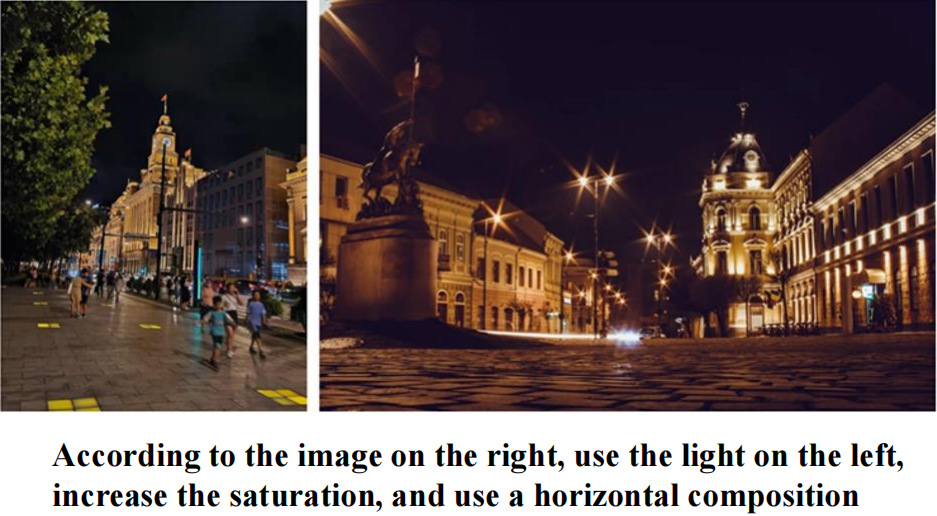}
	\caption{Night Landscape Aesthetics Guidance}
	\label{nightland}
\end{figure}

Now, we have an input image that needs guidance, and a guidance image that has been selected. Next, we apply the landscape aesthetic guidance described in the section 3 to the two images to determine their lighting direction, color palette, and composition type. If a certain attribute is different, for example, the input image is forward light, and the guidance image is left light, then the photographic scheme of the guidance image in lighting is using the left light; if a certain attribute of the two images is the same, use the aforementioned chapters, and output quantitative guidance results such as adjusting the degree of vividness and adjusting the camera angle. Figure \ref{nightland} is the final guidance result of Figure \ref{input}.
\begin{figure}
	\centering
	\includegraphics[height=3cm]{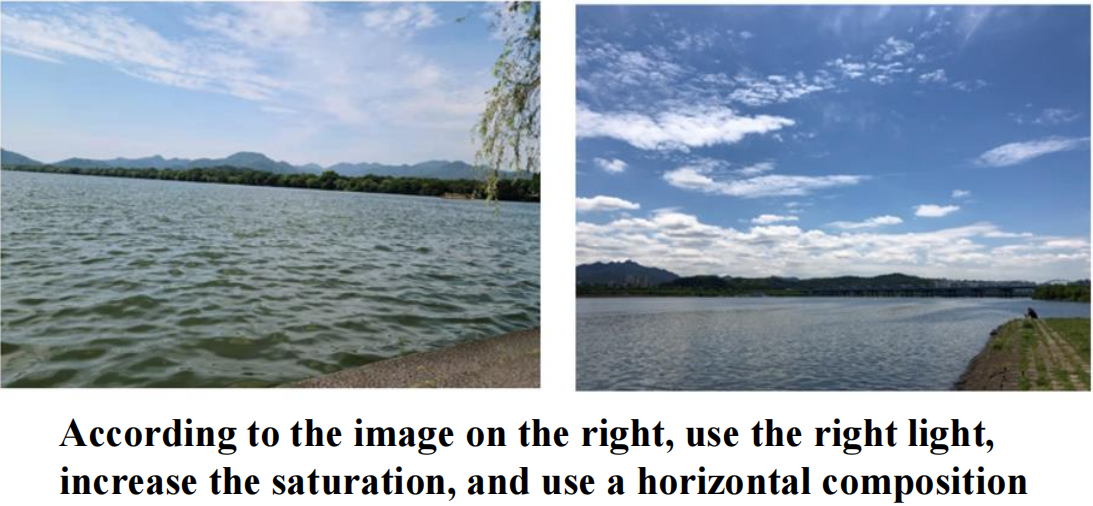}
	\caption{Daytime Landscape Aesthetics Guide}
	\label{Daytime}
\end{figure}

As shown in Figure \ref{Daytime}, we also present a photo of the aesthetic guidance results for daytime landscape photography.

\subsection{Aesthetic Portrait Images Guidance By ALG-I}
\begin{figure}[h!]
	\centering
	\includegraphics[height=4cm]{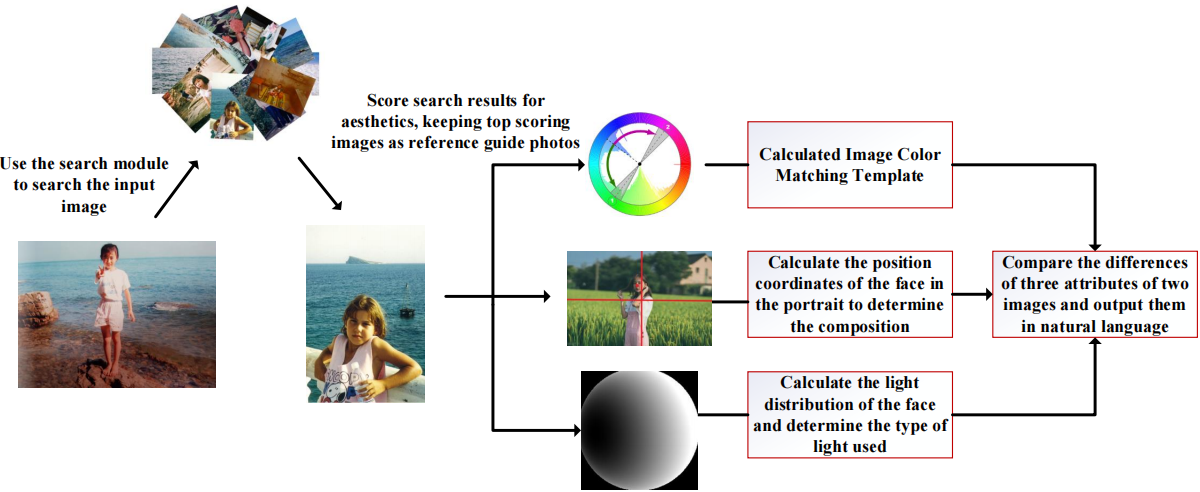}
	\caption{Portrait Aesthetic Image Guidance process}
	\label{porALGI}
\end{figure}

The basic idea of guidances for aesthetic portrait images is also that for any portrait image, using our image search system in the 4.1, we can search for images similar to the provided portrait image. It is then scored using a specific aesthetic scoring engine, with the highest scoring portrait image selected as the guidance image. The guiding rules for image color are the same as landscape images, but the composition type and lighting type are judged by the position of the face and the light distribution. Finally, compare the differences of the three attributes between the two images and output natural language for expression. The overall process is shown in Figure \ref{porALGI}.
\begin{figure}
	\centering
	\includegraphics[height=4cm]{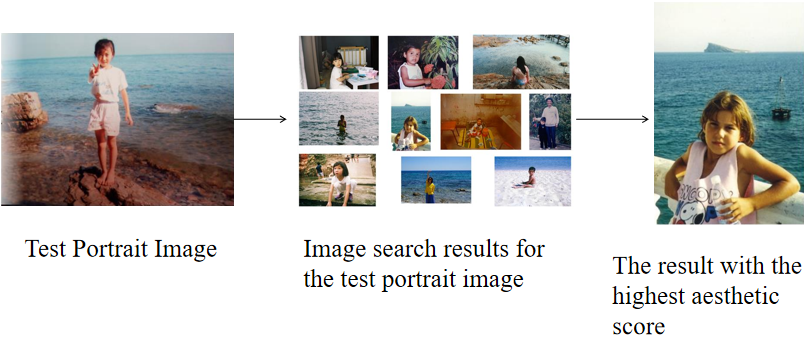}
	\caption{Image Search System for The Portrait Test Image}
	\label{portest}
\end{figure}
\begin{figure}
	\centering
	\includegraphics[height=3cm]{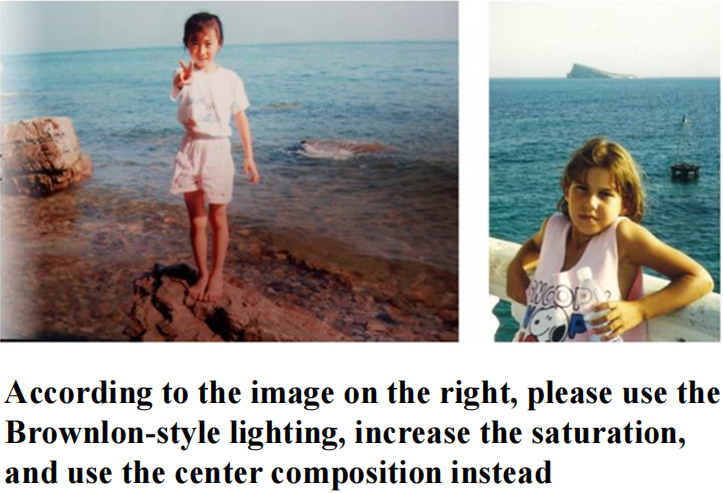}
	\caption{Aesthetic portrait guide by the sea}
	\label{latest}
\end{figure}

The overall process is similar to the section 4.2. The image on the left in Figure \ref{portest} is a portrait that needs guidance. The image in the middle of Figure \ref{portest} is the 10 candidate images most similar to the portrait test image obtained by the search module. Then use the aesthetic scoring engine to score them, and select the image with the highest score as the guidance portrait , as shown in the image on the right in Figure \ref{portest}. The differences between the portrait test image and the guidance portrait image are compared, and the results are shown in Figure \ref{latest}. 
\begin{figure}
	\centering
	\includegraphics[height=3cm]{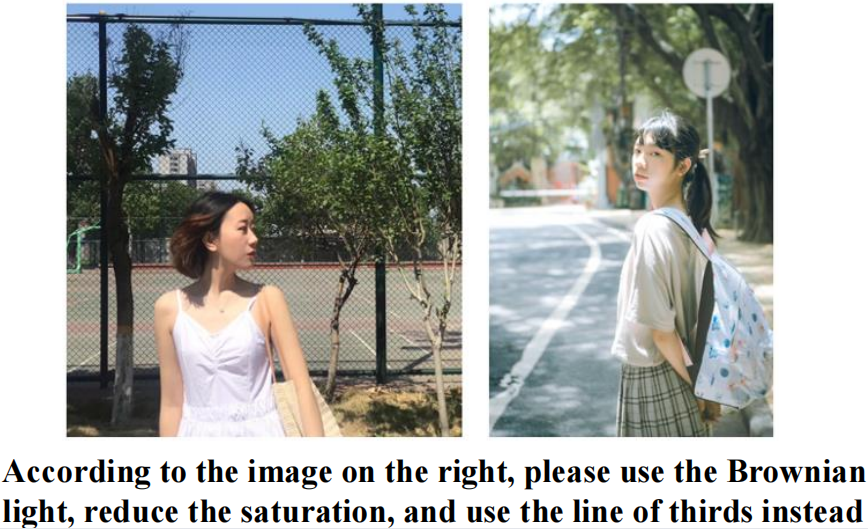}
	\caption{Aesthetic Guidance for Portraits in the City}
	\label{portraitcity}
\end{figure}

As shown in Figure \ref{portraitcity}, we also present a urban portrait photography aesthetic guidance result.

Whether it is a landscape photo or a portrait photo, it is necessary to search for similar images from the 2 million photographic image library based on the image search module, and evaluate the image aesthetic quality, and select the highest score as the guidance image for the input image. Then, according to whether it is a portrait image or a landscape image, different eigenvalues are calculated, different aesthetic standards are compiled, the differences between the attributes of the two are compared, and finally the guidance is expressed in natural language.

\section{Conclusions}
The work of this paper is mainly divided into two parts: the first part is language aesthetics shooting guidance based on photography templates. The second part is language aesthetics shooting guidance based on guidance images. Both of them make a distinction between portrait images and landscape images. And the color palette is calculated, the lighting distribution, semantic line classification, portrait composition, and face lighting are used as different guiding attributes. The guiding rules are designed and compiled, which are expressed in natural language.

On the other hand, there is no broad and unified definition and research scope for aesthetic guidance at present. The exploration in this paper belongs to only one family, and the detection indicators and guidelines are still superficial. In the future, more indicators can be added, more sophisticated data can be collected, more accurate guidance rules can be written, and the data flow between the running environment and the model can be unified, and the structure can be optimized on this basis to achieve real-time guidance close to the app's effects and abilities.

\bibliographystyle{splncs04}
\bibliography{egbib}

\end{document}